\documentclass[letterpaper]{article} 
\usepackage{aaai25}  
\usepackage{times}  
\usepackage{helvet}  
\usepackage{courier}  
\usepackage[hyphens]{url}  
\usepackage{graphicx} 
\usepackage{natbib}  
\usepackage{caption} 
\usepackage{algorithm}
\usepackage{algorithmic}

\usepackage{newfloat}
\usepackage{listings}
\DeclareCaptionStyle{ruled}{labelfont=normalfont,labelsep=colon,strut=off} 
\floatstyle{ruled}
\newfloat{listing}{tb}{lst}{}
\floatname{listing}{Listing}
\usepackage{booktabs}       
\usepackage{amsfonts,amsmath}       
\usepackage{subcaption}
\usepackage{adjustbox}
\usepackage{paralist}
\title{Differentiable Adversarial Attacks for Marked Temporal Point Processes}
\author{Pritish Chakraborty\equalcontrib\textsuperscript{1}, Vinayak Gupta\equalcontrib\textsuperscript{2}, Rahul R\textsuperscript{1}, Srikanta J. Bedathur\textsuperscript{3}, Abir De\textsuperscript{1}}
\affiliations{\textsuperscript{1}Indian Institute of Technology Bombay\\ \textsuperscript{2}University of Washington Seattle\\\textsuperscript{3}Indian Institute of Technology Delhi\\\{pritish, rrahul, abir\}@cse.iitb.ac.in, vinayak@cs.washington.edu, srikanta@cse.iitd.ac.in}

\usepackage{subdef}
\usepackage{xspace}
\usepackage{array}
\usepackage{multirow}
\newcommand{\our}{\textsc{PermTPP}\xspace}
\newcommand{\robtsd}{\textsc{RTS-D}\xspace}
\newcommand{\robtsp}{\textsc{RTS-P}\xspace}

\newcommand{\mifg}{MI-FGSM\xspace}
\newcommand{\mifgshort}{MIF\xspace}

\newcommand{\bestmodel}{\textbf}
\newcommand{\secondbest}{\underline}

\begin{document}

\maketitle

\begin{abstract}
Marked temporal point processes (MTPPs) have been shown to be extremely effective in modeling continuous time event sequences (CTESs).
In this work, we present adversarial attacks designed specifically for MTPP models. 
A key criterion for a good adversarial attack is its imperceptibility. For objects such as images or text, this is often achieved by bounding perturbation in some fixed $L_p$ norm-ball. However, similarly minimizing distance norms between two CTESs in the context of MTPPs is challenging due to their sequential nature and varying time-scales and lengths. We address this challenge by first permuting the events and then incorporating the additive noise to the arrival timestamps. However, the worst case optimization of such adversarial attacks is a hard combinatorial problem, requiring exploration across a permutation space that is factorially large in the length of the input sequence.  
As a result, we propose a novel differentiable scheme - \our\ - using which we can perform adversarial attacks
by learning to minimize the likelihood, while minimizing the distance between two CTESs.
Our experiments on four real-world datasets demonstrate the offensive and defensive capabilities, and lower inference times of \our.
\end{abstract}

%

\section{Introduction}

In recent years, marked temporal point processes (MTPPs) have become indispensable for modeling continuous time event sequences (CTESs). They find applications in various fields, including information diffusion~\cite{du2015dirichlet, de2016learning, farajtabar2017fake, sharma2021identifying}, finance~\cite{bacry2015hawkes, finance_hawkes,fullyneural,imtpp}, epidemiology~\cite{lorch2018stochastic, rizoiu2018sir}, and spatial data~\cite{colab, reformd}.
They offer impressive predictive power; nevertheless, our experiments reveal they are also vulnerable to adversarial attacks, similar to the neural models for images~\cite{trades,mart,madry,nuat} and texts~\cite{zou2023universal,jia2017adversarial}. Such vulnerabilities may compromise their reliability and robustness in several real world applications. For example, in information diffusion, they can jeopardize a viral marketing plan; in epidemiology,
they can affect policy implementation or epidemic control plans, etc.
Existing works on adversarial attacks predominantly focus on image~\cite{trades,rice2020overfitting,croce2020reliable, pmlr-v162-lin22e,huang2022transferable,wang2023structure,voracek2023improving,ma2023transferable} and texts~\cite{zou2023universal,jia2017adversarial,wei2024jailbroken,shin2020autoprompt,wen2024hard,jones2023automatically,zhao2024evaluating}, with few works on discrete time-series~\cite{oregi2018adversarial,karim2020adversarial,wu2022small,fawaz2019adversarial,ma2020adversarial,yoon2022robust,ding2023black,belkhouja2022dynamic}. However, there is still a lack of research on designing adversarial attacks for MTPPs, which if unaddressed, may leave potential vulnerabilities in the current deployed MTPP-based systems~\cite{tpp_review}.

\xhdr{Challenges in designing attacks for MTPPs}
In this work, we focus on developing adversarial attacks specifically for MTPP models, which can help design more robust training methods. 
MTPPs are very different from images in terms of both representations and modeling characterizations. 
Hence, the existing adversarial attacks designed for images are not suitable for MTPPs. 
Moreover, due to the disparate inter-event arrival times, MTPPs pose the following challenges in designing both adversarial attacks and their defense.

\ehdr{Order-induced complexity of perturbation} 
In the context of images, measuring adversarial perturbation between clean and perturbed objects is relatively straightforward. The Euclidean distance between pixel matrices serves as a suitable distance metric, generating zero-mean i.i.d. additive adversarial perturbations for each pixel value. The strength of the perturbation is quantified 
using the maximum difference between corresponding pixel values.

The concept of adversarial perturbation becomes more complex in the context of continuous time event sequences (CTESs). In contrast to images, the input data consists of an ordered set of events. Here, introducing noise to the arrival times may not manifest a simple monotone relationship with the overall perturbation of the sequence. For instance, applying a high level of noise to each event may still yield a smaller overall perturbation. This is because the perturbed sequence closely resembles the clean sequence offset with nearly constant noise, thereby maintaining the order of events unchanged.
In such cases, the lack of impact on the temporal ordering weakens the overall attack. 
On the contrary, a small amount of noise might lead to a larger overall perturbation as it alters the order of events.   

 
\ehdr{Non-differentiability in adversarial generation} 
In the context of MTPP, even a small temporal noise can change the event order. Such a change causes an abrupt shift in the embedding vector within the MTPP model, which introduces non-differentiability in the overall model. This hinders the model's ability to accurately learn from perturbed data.

\subsection{Our Contributions}

We present \our, a novel trainable adversarial attack for CTES. \our handles perturbations in event ordering, event marks, and event times, \ie, all associated properties of an MTPP. In detail, our contributions are:

\xhdr{Novel formulation of adversarial attacks} 
Given a CTES $\Hcal$, we generate a perturbed sequence $\Hcal'$ through a two-stage adversarial attack, followed by worst case likelihood minimization.
In the first stage, we adversarially permute the events, while explicitly controlling the extent of shuffling. To do so,  we permute the events in such a way that the total deviation in event positions from the clean sequence is small. 
In the second stage,  we add adversarial noise to the arrival time of each event in the permuted sequence. Finally, we compute the perturbed sequence by minimizing the worst case likelihood over all possible permutations with additive noise, subject to the constraint that distance between the original and the perturbed sequence remains small.   

\xhdr{Differentiable permutation guided event reordering}
Worst case likelihood optimization over the set of all possible permutations is a hard combinatorial challenge.  To tackle this challenge, we generate a soft permutation matrix through a differentiable permutation scheme, built upon a Gumbel-Sinkhorn (GS) network~\cite{mena2018learning}.  Based on the MTPP model available to the adversary, GS takes the embeddings of the model as input and outputs a doubly stochastic matrix, which serves as a differentiable surrogate for the hard permutations.

\xhdr{Order-constrained additive noise injection}
Having permuted the sequence (both time and mark) as above, we introduce a temporal noise to each of the event timestamps, through a noise generator. However, this additive noise might further shuffle the ordering of the events, which is already decided based on the event reordering in the previous step. We prevent such further event shuffling, by adding a hinge loss regularizer into the attacker's overall loss.

Our experiments on four real-world datasets show that \our\ is able to significantly degrade an MTPP model, trained on both clean and adversarially perturbed CTESs.

\section{Problem Formulation}
\xhdr{Marked temporal point processes (MTPPs)} MTPPs model sequences of discrete events in continuous time.   An event $e = (t,c)$ where $t \in \RR^+$ and $c\in \Ccal$ denote the arrival time and the mark, respectively, with $\Ccal$ being discrete.
Thus, CTES can be seen as $\Hcal = \{e_i=(t_i,c_i) \given  t_{i-1}<t_{i}\}$, where $e_i$ is the $i$-the event in the CTES. We denote $\Hcal(t)$, as the history of events occurred until and excluding time $t$, and a counting process $N(t)$, which counts the number of events that occurred until time $t$. Then, we define the rate of events using the conditional intensity function $\lambda(t \given \Hcal_t)$, which is computed as:
$  \EE[dN(t)\given \Hcal_t] = \lambda(t \given \Hcal_t) dt.
    $
Here, $dN(t)\in\set{0,1}$ is number of events arriving in the infinitesimal time interval $[t,t+dt)$, with $\lambda(t \given \Hcal_t)$ depends on history. We model the mark distribution using multinomial distribution $c\sim \markx(\bullet \given \Hcal_t)$, which is used to draw the mark of the event arriving at time $t$. 
Given $\Hcal(t_i)$,  the expected time of the \textit{next} event is:
$    \EE \left[ t_{i+1} |\Hcal(t_i) \right ] = \int_{t_i}^{\infty} t \cdot \lambda(t \given \Hcal_t) \, dt.
$
Moreover, we sample the mark of the next event from the discrete distribution over all possible categories.

The task of designing MTPPs reduces to developing appropriate $\lambda_{\wb}(t\given \Hcal_t)$ and $\markx_{\wb}(\bullet \given \Hcal_t)$ for the intensity and mark distribution.
~\citet{du2016recurrent,MeiE16} use recurrent networks, whereas, 
~\citet{zuo2020transformer} and~\citet{zhang2019self}
utilize the self-attention to capture long-term dependencies.
In our experiments, we use the transformer-based MTPP model~\cite{zuo2020transformer}. 


\xhdr{Additional notations} Given an integer $I$, we denote $\pi: [I] \to [I]$ as a permutation of the integers $\set{1,...,I}$ and $\Pb$ as the corresponding permutation matrix of dimension $I\times I$. Given a neural MTPP model, we will frequently use $\hb_i$ as the embedding of
$\Hcal(t_i)$ until time $t_i$. 
Typically, they summarize the first $i$ events into  $\hb_i$ using a sequence encoder and feed this as input into a final layer (sometimes, with the event $e_{i}$) which output the intensity and the logit for the mark distribution.  Hence, one can equivalently write:
$\lambda_{\wb}(t_{i+1} \given \Hcal_{t}) = \lambda_{\wb}(t_{i+1} \given \hb_i)$ and $\markx_{\wb}(c_{i+1} \given \Hcal_t) = \markx_{\wb}(c_{i+1} \given \hb_i)$. We denote $\lmodel$ as the learner's trained model and $\amodel$ as the adversary's model--- or, the adversary's belief about the learner's trained model.


\subsection{Problem Statement}
Here, we formulate the optimization problem for adversarial attacks and their defense mechanisms on MTPPs.

 
 \xhdr{Adversarial attack for MTPP} Given an MTPP model $\amodel$ with parameters $\wb$ with the adversary,  we seek to develop methods for adversarial attacks, which would result in significant error in terms of 
both time and mark prediction.  Here the model $\amodel$ may not match with the learner's true model and it is rather the adversary's belief about the learner's true model. 
Given a clean sequence $\Hcal$,  the adversary seeks to find the adversarial sequence $\Hcal' = \set{(t' _i, c' _i) \given t' _i < t' _{i+1}, i\in [|\Hcal|]}$\footnote{\scriptsize $\set{(t' _i, c' _i)}$ is always chronologically sorted: $t' _{i+1} > t' _i$. Hence, $t' _i$ is not necessarily the perturbed time from $t _i$.}
so that the adversary's model $\amodel$ incurs high error in predicting the correct event $(t_i,c_i)$, when it takes $\Hcal'(t_i)$ as input.
Thus, we define the adversary's objective function as follows.
\begin{align}
   &\Lcal (\Hcal\given \Mcal_{\wb},\Hcal')  
    =\sum_{e_i \in \Hcal }\bigg[\log \lambda_{\wb}(t_i \given \Hcal' (t_i)) \nn\\
    &-  \int_{t' _{i-1}} ^{t_i} \lambda (\tau \given \Hcal' (\tau)) d\tau  + \log \markx_{\wb} (c_i \given \Hcal' (t_i)) \bigg] \label{eq:obj0} 
\end{align}  
Then, the adversary has the following optimization problem.
\begin{align}
&\underset{\Hcal'}{\mini}\,  \Lcal (\Hcal \given \Mcal_{\wb},\Hcal')\label{eq:attack0} \, \, \text{s.t., }\  \dist(\Hcal, \Hcal') \text{ is small}.
\end{align}
Hence, the adversary seeks to minimize the likelihood of the \emph{clean events} $e_i$, conditioned on the perturbed events $\Hcal'(t_i)$ occurred until $t_i$.
Note that the second term in ~\eqref{eq:attack0} is slightly different than  the usual survival term in the clean likelihood $(\int_{\bm{t' _{i-1}}} ^{t_i} \hspace{-2mm}\lambda (\tau\given \Hcal (\tau)) d\tau  $ vs. $ \int_{\bm{t _{i-1}}} ^{t_i} \hspace{-2mm}\lambda (\tau\given \Hcal (\tau)) d\tau)$.

\subsection{Computation of $\dist(\Hcal,\Hcal')$}
  Suppose, we are given a sequence of events $\set{(t_i + \epsilon_i, c _i)}$  after adding noise $\epsilon_i$ to the timestamps. Since $\epsilon_i$ are not uniform across $i\in[|\Hcal|]$, the resulting sequence is not sorted. Hence, we first sort $\set{(t_i + \epsilon_i, c _i)}$ using the following:
\begin{align}
 & \pi = \text{argsort}(\set{t_i\hspace{-1mm}+\noise_i}_{i\in[|\Hcal|]}) , t' _{1} =\text{min}(\set{t_i\hspace{-1mm}+\noise_i}_{i\in[|\Hcal|]})\nn 
 \\ \label{eq:argsort}
 &\quad \implies  \Hcal' = \set{(t' _i, c' _i) \given t' _i= t_{\pi_i}, c' _i=c_{\pi_i}}
\end{align}
Therefore, $\pi$ indicates the permutation of the indices after sorting the events in the increasing order of the perturbed arrival times.
In most real world scenarios, MTPP datasets do not provide information about the time  $t_{\text{start}}$ from which the event monitoring started. The time $t_{\text{start}}$ is different from the first arrival time $t_1$ for $\Hcal$ (or $t_1'$ for $\Hcal'$), since the latter is recorded only \emph{after} we start the monitoring. An MTPP model assumes some initial monitoring time, mostly either $t_{\text{start}}=0$ or the time of the first event, \ie,  $t_{\text{start}}=t_1$. If we choose a fixed $t_{\text{start}}$, agnostic to the underlying sequence (\eg, $t_1=0$), then  the distance between $\Hcal'$ and $\Hcal$ becomes very sensitive to $t_{\text{start}}$. Therefore, we use $t_{\text{start}}=t_1$ and $t_{\text{start}}=t_1 '$ for $\Hcal$ and $\Hcal'$. This allows us to offset the arrival times of  $\Hcal$ and $\Hcal'$ from the respective initial times $t_1$ and $t_1'$ and then measure the Wasserstein distance, $\dist(\Hcal, \Hcal')$, as follows~\cite{xiao2017wasserstein}.
\begin{align}
   \hspace{-1mm}  \dist(\Hcal, \Hcal')  = \hspace{-1mm}\sum_{i\in[|\Hcal|]} \Big[|(t' _{i}    -t'_{1}) - (t_i-t_1)| + \rho_c\II[c' _{\pi_i} \neq c_i]\Big],\hspace{-1mm} \label{eq:dd} 
\end{align}
Here, $\rho_c$ controls the contribution between time and mark. 

\subsection{Key Technical Challenges} \label{prob:challenges}

\xhdr{Difficulty in generating controlled perturbation $\Hcal'$} A fundamental constraint of the adversarial perturbation is that 
 the amount of perturbation, \ie, the distance between $\Hcal$ and $\Hcal'$,  has to be small enough to be undetectable.
In the context of images or multivariate time series~\cite{goodfellow2014explaining}, the amount of perturbation simply corresponds to the difference between the underlying feature vectors. 
Consequently, in these applications, the task of generating adversarial objects is straightforward, as it involves adding controlled magnitude adversarial noise to the existing data. Unlike images, simply controlling the magnitude of the noise may not control $\dist(\Hcal,\Hcal')$. This is because the value of $\noise_i$ has a non-monotone relationship with $\dist(\Hcal, \Hcal')$, which manifests through the argsort operation.  
\begin{figure}
 \centering
\includegraphics[width=0.40\textwidth]{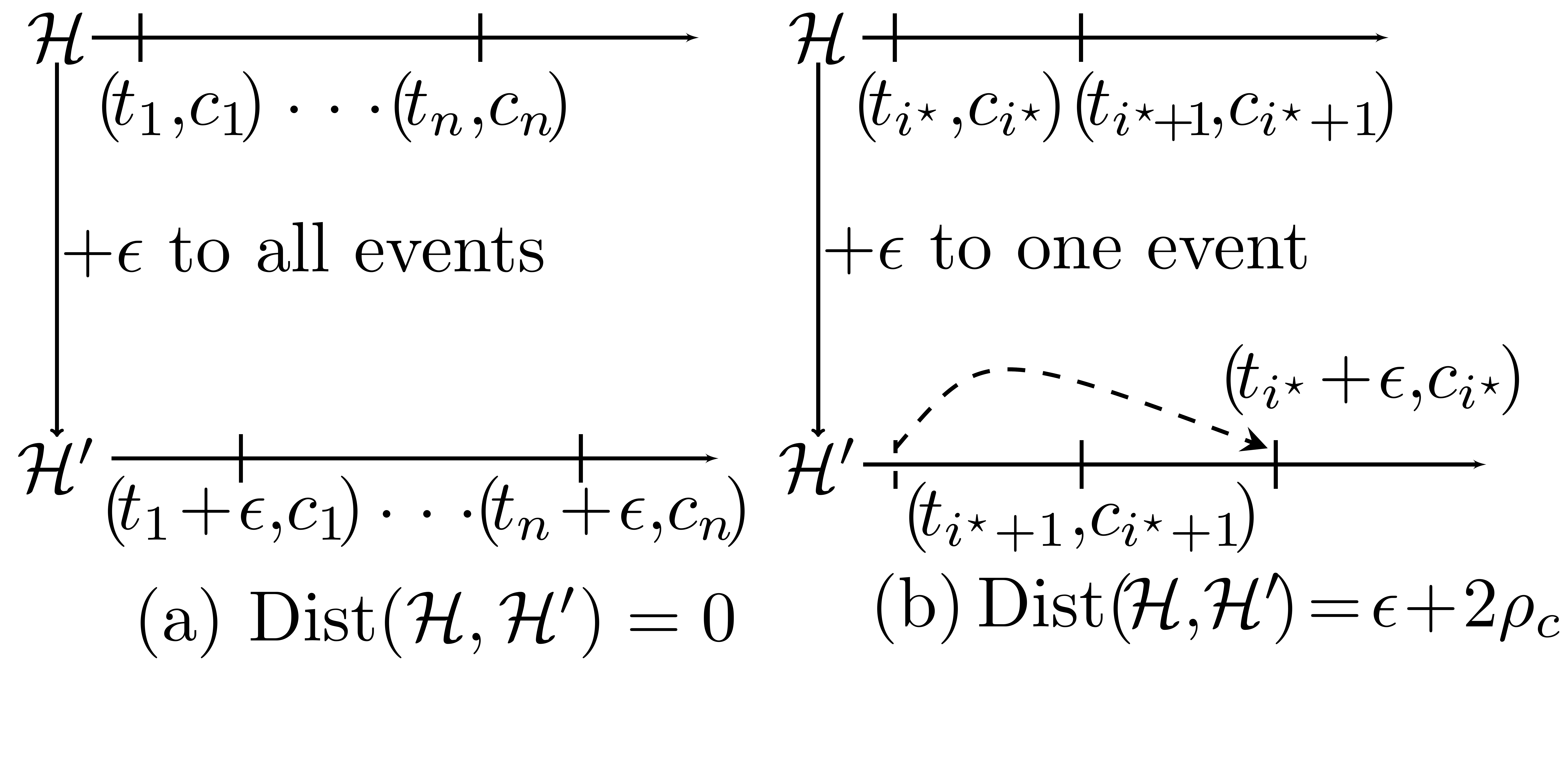}
    \caption{\small More noise may not increase $\dist (\Hcal,\Hcal')$.}
 \label{fig:Dist_demo}
\end{figure}

For instance, consider $c'_i = c_i$, $\epsilon_i=\epsilon$.  Despite the significant amount of perturbation introduced in $\Hcal'$, both $\Hcal'$ and $\Hcal$ share the same order of events. Consequently, the distance between the mark sequences becomes zero. Moreover, we have $(t_{i} + \epsilon _{i} -t' _{\min}) - (t_i -t_1)=0$ for all $i$. Hence, $\dist (\Hcal,\Hcal')= 0$ (Figure~\ref{fig:Dist_demo} (a)). 
On the contrary,
we can construct $\Hcal'$ by adding a small amount of noise to a single event, which swaps it with its successive event.
Specifically, assume that for some $i=i^{\star}$,  we have $c_{i^{\star}}\neq c_{i^{\star}+1}$. We choose $\epsilon_{i^{\star}}=\epsilon$ such that $t_{i^{\star}+1}-t_{i^{\star}}<\epsilon< t_{i^{\star}+2}-t_{i^{\star}}$ and add it to $t_{i^{\star}}$ (Figure~\ref{fig:Dist_demo} (b)). For all $i\neq i^{\star}$, we have $\epsilon_i =0$.  This leads $e_{i^{\star}}$ and $e_{i^{\star}+1}$ exchanging their positions in the resulting sequence. Moreover, we have
$t'_1=t_1< t' _2 = t_2<...<t' _{i^{\star}} = t _{i^{\star}+1}<t'_{i^{\star}+1}=t_{i^{\star}} + \epsilon < t' _{i^{\star}+2} = t_{i^{\star}+2}<...<t'_n =t_n$, resulting in perturbation of the event ordering. A trivial calculation shows that: 
\begin{align}
\dist(\Hcal,\Hcal') & = 
\textstyle\sum_{i\in[|\Hcal|]} \Big[|(t' _{i}  -t'_{1}) - (t_i-t_1)| \\
& + \rho_c\II[c' _{\pi_i} \neq c_i]\Big]\nn\\
& =|t_{{i^{\star}}+1}  -t_{{i^{\star}}}| + |t_ {{i^{\star}}} +\epsilon  -t_{{i^{\star}}+1}|    \\
&+\rho_c \II[c_{{i^{\star}}+1} \neq c_{i^{\star}}] + \rho_c\II[c_{{i^{\star}}} \neq c_{{i^{\star}}+1}]\nn\\
&= (t_{{i^{\star}}+1}  -t_{{i^{\star}}})+\epsilon-(t_{{i^{\star}}+1}  -t_{{i^{\star}}})+2\rho_c \nn\\
&=\epsilon+2\rho_c.      
\end{align}
Hence, we have a non-zero distance between the both time and mark sequences. 
Thus, using additive noise to control the distance between $\Hcal$ and $\Hcal'$ is a challenging task.

\xhdr{Non-differentiability of adversarial loss} 
The objective of adversarial perturbation is to induce a significant prediction error, which leads to intrinsic non-differentiability in the perturbed model. However, in other applications, this effect is less pronounced in the initial layers of the model due to the smoothness in the feature perturbation process--- a small-magnitude noise causes only minor perturbations of the initial features. However, such a smoothness does not apply to perturbations on CTESs, resulting in increased non-differentiability for perturbed MTPP models.

To illustrate this point, consider the previous example where $\Hcal' = \set{(t_i' = t_i + \epsilon_i, c_i)}$ and $\epsilon_{i^{\star}} >  t_{{i^{\star}}+1} - t_{i^{\star}}$ for some  $i={i^{\star}}$, and $\epsilon_i = 0$ for all $i \neq {i^{\star}}$. In this scenario, due to the shift in event ordering, $\Hcal'$ significantly differs from the original $\Hcal$, even with the addition of small noise. Consequently, this leads to an abrupt jump in the embedding $\hb_{i^{\star}}'$, resulting in a high value of $||\hb_{i^{\star}} - \hb_{i^{\star}}'||$.

\section{\our\ Model}
In this section, we describe \our, our proposed adversarial attack model and its different components. 


\subsection{Proposed Adversarial Attack Model}
\xhdr{Overview of our method} The key hurdle behind controlling the extent of perturbation in the sequence $\Hcal'$ is the involvement of the argsort operation in distance computation in Eq.~\eqref{eq:argsort}, which changes the event ordering. To tackle this problem we first decide on $\pi$, the chronology of the events before adding the underlying noise. Then, we add the noise $\set{\noise_i}$ in an adversarial manner, subject to the constraint that the resulting chronological order after adding the noise does not change from $\pi$. In the following, we build \our, Permutation Guided Adversarial Attack in two stages.

\xhdr{Combinatorial formulation of two staged adversarial attack}
We note that the difference between two CTESs $\Hcal$ and $\Hcal'$ is guided by three factors: (i) the discordance between the ordering of the events; (ii) the difference between the arrival times; and (iii) the mismatch between the marks. 
Therefore, we generate adversarial perturbations in two stages.
First, we perform an adversarial permutation on the events to generate a new perturbed order. This corresponds to factors (i) and (iii) above. Then, we add suitable noise to the timestamps of those events, subject to the above order. This corresponds to factor (ii).

\ehdr{Adversarial permutation} 
Consider a clean CTES $\Hcal$, which consists of $n$ events. In the first stage, we explicitly apply a permutation $\pi: [n] \to [n]$ on $\Hcal$ before adding any noise. This results in a new permuted sequence denoted as $\Hcal_{\pi} =\set{(t_{\pi_1},c_{\pi_1}),(t_{\pi_2}, c_{\pi_2}), \ldots, (t_{\pi_n},c_{\pi_n})}$.
Note that unlike $\Hcal$, the ordered set $\Hcal_{\pi}$ may not maintain a chronological order, and thus, the condition $t_{\pi_i} < t_{\pi_{i+1}}$ may not hold.

\ehdr{Introducing temporal noise} 
In the second stage, we proceed to add adversarial noise $\set{\noise_i}_{i \in [n]}$ to the timestamps of the permuted sequence $\Hcal_{\pi}$. This results in a perturbed sequence $\Hcal_{\pi,\bm{\noise}} = \set{(t'_i,c' _i)}$, where $t'_i = t_{\pi_i} + \noise_i$ and $c' _i = c_{\pi_i}$. We select the additive noise $\noise_i$ in a manner that the perturbed times become chronologically ordered and aligned with the permutation $\pi$, which was already computed in the first stage. This allows imposing constraints on $\noise_i$ such that:
\begin{align}
&t_{\pi_i} + \noise_i < t_{\pi_{i+1}} + \noise_{i+1} \text{ for all } i < n, \quad 
t_{\pi_1} + \noise_1 > 0. \label{eq:tD1}
\end{align}
Let us define $\bm{\noise} = [\noise_1; \ldots; \noise_n]$ and $\bm{t}_{\pi} = [t_{\pi_1}; \ldots; t_{\pi_n}]$. 
Constraints in Eq.~\eqref{eq:tD1} are linear constraints with respect to $ \bm{t}_{\pi}$ and $\bm{\noise}$. Therefore, using suitable coefficient matrices $\Ab$ and $\Bb$,  they can be re-written as follows:%
\begin{align}
\Ab \bm{\noise} < \Bb \bm{t}_{\pi} \label{eq:AB}
\end{align}

Here, $\Ab, \Bb \in \mathbb{R}^{n\times n}$, with $\Ab[i, i]=1$, $\Ab[i,i+1]=-1$,  $\Bb[i, i]=-1$, $\Bb[i,i+1]=1$ for $i\in [n-1]$ and $\Ab[n, 1]=-1$, $\Ab[n,j]=0$, $\Bb[n, 1]=1$, $\Bb[n,j]=0$ for $1<j\leq n$. Additionally, $\Ab[i, j] = 0, \Bb[i, j] = 0\ \forall j \neq i$ and $j \neq i + 1, i \in [n - 1]$.
This formulation allows us to ensure that the perturbed timestamps maintain the desired chronological order, aligning with the predefined permutation $\pi$ obtained in the first stage. Hence, we have our adversarial sequence as $\Hcal' = \Hcal_{\pi,\bm{\noise}} = \set{(t' _i, c'_i ) \given t'_{i} < t' _{i+1} }$ where $t_i ' = t _{\pi _i} + \epsilon_i$. 

\xhdr{Optimization of $\pi,\epb$} 
We first re-write the 
distance measure $\dist(\Hcal,\Hcal_{\pi,\bm{\noise} })$~\eqref{eq:dd} as follows:
\begin{align}
\dist(\Hcal,\Hcal_{\pi,\bm{\noise}}) = & \textstyle\sum_{i\in[|\Hcal|]} |t_{\pi_i} \hspace{-1mm}+  \epsilon_{\pi_i} \hspace{-1mm} - t_i|    \nn \\ & + \rho_c\, \left(1-m_{\wb}(c'_{\pi_i}=c _i \given \Hcal_{\pi,\epb} )  \right) \label{eq:distDef}
\end{align}
Here
the first term measures the deviation due to adversarial permutation the additive noise and 
the second term measures the expected number of times for which the adversarially generated mark is different than the true mark, \ie, $\sum_{i}\EE[\II[c'_{i}\neq c_i]]$ where $\II[\cdot]$ is the indicator function.  Given the above measure of the distance in Eq.~\eqref{eq:distDef} and constraints~\eqref{eq:AB}, we write the problem defined in Eq.~\eqref{eq:attack0} 
as a combinatorial optimization task, with respect to the combinatorial variable $\pi$ and the continuous variable $\epb$ as follows:
\begin{align}
   &\underset{\pi,\bm{\noise}}{\mini} \   \Lcal (\Hcal \given \Mcal_{\wb},\Hcal_{\pi,\bm{\noise}}) + \rho_{D}  \dist(\Hcal,\Hcal_{\pi,\bm{\noise}}) \nn \\
   &\text{subject to,} \  \Ab \bm{\noise} \le \Bb \tb_{\pi} \label{eq:m0} 
\end{align}
We encode the constraint $\Ab \bm{\noise} \le \Bb \tb_{\pi}$ into a hinge loss
$\textstyle\sum_{i\in[|\Hcal|]}[\Ab \bm{\noise} -\Bb \tb_{\pi}]_+[i]$  and add it to the objective~\eqref{eq:m0} to solve the following regularized optmization task, with hyperparameters $\rho_{D}$ and $\rho_{A,B}$
\begin{align}
  & \hspace{-2mm} \underset{\pi,\bm{\noise}}{\mini} \   \Lcal (\Hcal \given \Mcal_{\wb},\Hcal_{\pi,\bm{\noise}}) + \rho_{D}  \dist(\Hcal,\Hcal_{\pi,\bm{\noise}})\nonumber\\
  &+ \rho_{A,B}  \textstyle\sum_{i\in[|\Hcal|]}[\Ab \bm{\noise} -\Bb \tb_{\pi}]_+[i]  \label{eq:attack_final}
\end{align}


\subsection{Differentiable Neural Networks for $\pi, \noise$}

\begin{figure*}[t]
\centering
\includegraphics[width= 0.9\textwidth]{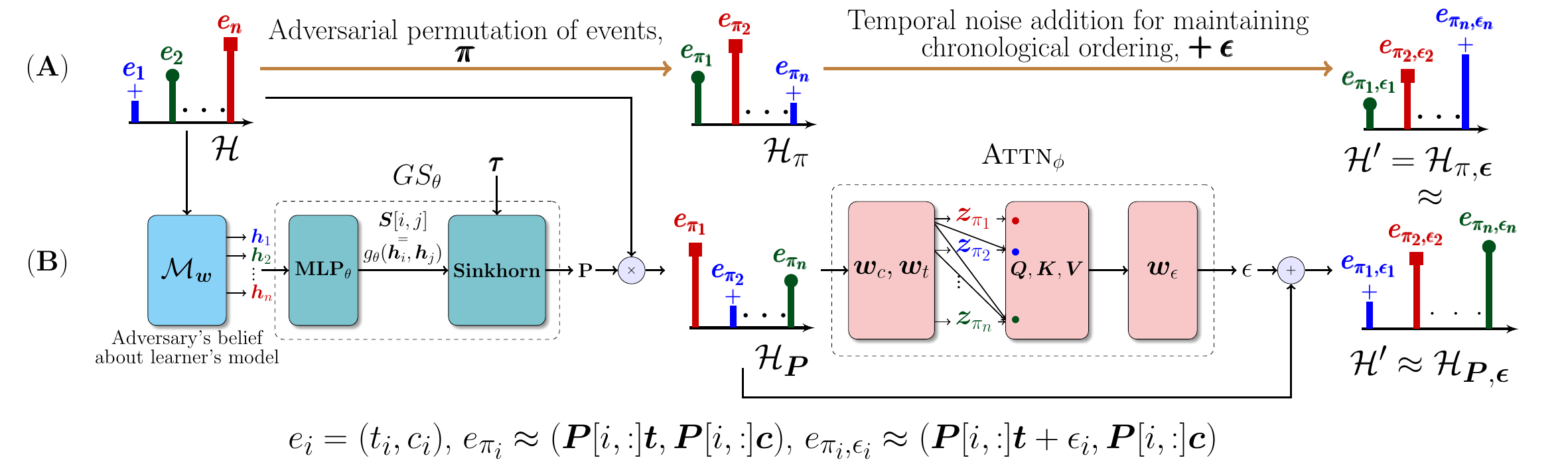}
\caption{Overview of \our. \textbf{(A) Combinatorial perspective of two staged adversarial attack}: It describes combinatorial computation of the adversarial CTES $\Hcal'=\Hcal_{\pi,\epb}$ from 
the clean CTES $\Hcal$.
We first apply a (hard) permutation map $\pi$ on  $\Hcal$ to obtain the permuted sequence $\Hcal_{\pi}$.
The map $\pi$ was intended to fix the chronological order of the events of the final perturbed CTES $\Hcal'$. However $\Hcal_\pi$ is not chronologically sorted. 
Hence, we add temporal noise $\bm{\noise}=\set{\epsilon_1,..\epsilon_n}$ to the timestamps in $\Hcal_\pi$ with a hinge regularizer (last term in Eq.~\eqref{eq:attack_final}),
so that $\Hcal'=\Hcal_{\pi,\bm{\noise}}$ is chronologically ordered and is aligned w.r.t $\pi$.    \textbf{(B) Differentiable approximation of $\pi,\epb$}: The CTES $\Hcal=\set{e_i \given i\in[n]}$ is fed to $\Mcal_{\wb}$ (adversary's belief about the learner's model $\lmodel$) to obtain the embeddings $\Hb=\set{\hb_i\given i\in [n]}$. They are fed to a Gumbel-Sinkhorn network $GS_{\theta}$ to obtain a soft permutation matrix $\Pb$. Within $GS_{\theta}$, first an MLP network $g_\theta$ generates the seed matrix $\Sb$ with $\Sb[i,j] = g_{\theta}(\hb_i,\hb_j)$ and then $\Sb$ goes through a number of Sinkhorn iterations with temperature $\tau$, to output $\Pb$; $\Pb$ is then applied on $\Hcal$ to obtain the soft permuted sequence  $\Hcal_{\Pb}\approx\Hcal_{\pi}$. Next, $\Hcal_{\Pb}$ is fed into $\epsAttn$, which consists of of linear, attention and linear layers, to obtain the additive noise $\epb$, which finally provides $\Hcal_{\Pb,\epb}\approx \Hcal_{\pi,\epb}$.}  
\label{fig:test}
\end{figure*}

Solving the optimization problem defined in Eq.~\eqref{eq:attack_final} is hard due to the large number of optimization variables of order $O(\sum_{\Hcal\in \Dcal}|\Hcal|)$ as well as the involvement of the combinatorial variables $\set{\pi}$. To address this challenge, we develop a neural network $\our_{\theta,\phi}$ which would model $\pi$ and $\bm{\epsilon}$, based on $\Hcal$ used as input to the neural network. Training this network to solve the optimization~\eqref{eq:attack_final} will provide the necessary inductive bias into $\our_{\theta,\phi}$ to compute the optimal $\Hcal'$ for an unseen $\Hcal$.
Thus, $\our_{\theta,\phi}$ has two components, as follows: \\
\textbf{(1) $GS_{\theta}$: } We feed the embeddings $\set{\hb_i}$ from $\amodel$, that summarizes the sequence $\Hcal$ into a Gumbel-Sinkhorn permutation generator $GS_{\theta}$ and outputs $\Pb$, a soft permutation matrix, which \smash{relaxes $\pi$}.\\
\textbf{(2)}  $\epsAttn$:  This network takes $\Pb$ and $\Hcal$ as input and outputs the additive noise $\epb$.

\xhdr{Neural model for $\pi$ ($GS_{\theta}$)}
For batched learning, we appropriately pad each sequence $\Hcal\in \Dcal$ to obtain the same length $n$. 
One can replace $\pi$ with a hard permutation matrix $\Pb$ of dimensions $n\times n$, where $\Pb[i,j]=1$ iff $j=\pi[i]$. To ensure end-to-end continuous optimization, we relax this hard permutation matrix using a soft permutation matrix (a doubly stochastic matrix), which is learned using
Gumbel-Sinkhorn~\cite{mena2018learning} network $GS_{\theta}$.
The neural TPP models~\cite{du2016recurrent,MeiE16} provide embeddings
$\hb_i$, for each event $e_i\in \Hcal$, which represent the history $\Hcal(t_i)$ upto time $t_i$. Given the TPP model $\Mcal_{\wb}$ available to the adversary (\ie, adversary's belief about the learner's model $\lmodel$), we feed these embeddings $\Hb=\set{\hb_i\given i\in [n]}$ into the Gumbel-Sinkhorn permutation generator $GS_{\theta}$ to generate the soft permutation matrix $\Pb$.
We write that:
\begin{align}
    \Pb = GS_{\theta} (\Hb), \quad \text{where: }\Hb=\set{\hb_i\given i\in [n]} \label{eq:PH}
\end{align}
$GS_{\theta}$ consists of a multi-layer perceptron (MLP) $g_{\theta}$ and a Sinkhorn iteration layer.
We feed $\Hb$~\eqref{eq:PH} into $g_{\theta}$ to obtain a matrix $\Sb$ such that $\Sb[i,j] = g_{\theta}(\hb_i,\hb_j)$. This matrix $\Sb$ works as a seed matrix to generate $\Pb$ through successive Gumbel-Sinkhorn iterations from $\iter=1,...,\itermax$. Specifically,
we start with $\Ab_0$ as
$    \Ab_0 = \exp(\Sb/\tau)$.
Then, given $\Ab_{\iter-1}$ computed after iteration $\iter-1$ for $\iter>0$, we compute $\Ab_{\iter}$. For this, we normalize each row of $\Ab_{\iter-1}$ to obtain $\Ab^{(r)}  _\iter$ and then we normalize each column of $\Ab^{(r)}_l$ to obtain $\Ab_\iter$. 
\begin{align}
    &\Ab^{(r) } _{\iter} [i,j] =\frac{ \Ab_{\iter-1}[i,j] } {\sum_{j'=1} ^n \Ab _{\iter-1} [i,j']}; \,  
    \Ab _{\iter} [i,j] = \frac{   \Ab^{(r) } _{\iter} [i,j]}{  \sum_{i'=1} ^n \Ab^{(r) } _{\iter} [i',j]} \nn \\
    & \hspace{3cm} \text{for all} \ i \in [n], j \in [n] \label{eq:Abc}
\end{align} 
Then we set $\Pb=\Ab _L$. Note that, $\Pb$ also approximately solves the following optimization problem:
\begin{align}
&\underset{\Pb}{\max}\     \sum_{i,j}\Sb[i,j]\Pb[i,j] - \tau \cdot \sum_{i,j} \Pb[i,j] \log \Pb[i,j],\\
&\hspace{2cm} \text{such that } \Pb \ge 0, \ \Pb \bm{1}=\Pb^{\top} \bm{1} = \bm{1}.\nonumber  
\end{align} 
As $\tau \to 0$, $\Pb$ becomes close to hard permutation matrix~\cite{mena2018learning,cuturi}. However, for a very low value of $\tau$, the gradient becomes almost zero. As a result, the learner cannot leverage any feedback from the loss. 
Finally, we apply $\Pb$ on the $\Hcal$ to obtain the new (soft) permuted sequence as: $\Hcal_{\Pb} = [\Pb \tb, \Pb \bm{c}]$, where $\tb=[t_1,..,t_n]$ and $\bm{c}=[c_1,...,c_n]$.

\xhdr{Neural model for $\epb$ ($\epsAttn$)}  
The next stage of attack involves generating continuous noise to add to the permuted sequence $\mathcal{H}_{\Pb} = [\Pb \tb, \Pb \bm{c}]  $. The continuous noise is constrained by Eq~\eqref{eq:AB} to ensure that order of the events remain the same as decided by $\Pb$ computed as above in Eq.~\eqref{eq:Abc}. To determine the noise for each event $(t_{\pi_i}, c_{\pi_i})\approx (\Pb[i,:]\tb,\Pb[i,:]\bm{c})$, we employ an attention based network $\epsAttn$, which takes the permuted mark and time sequences, \ie\ $\mathcal{H}_{\Pb}$ as inputs, and outputs an intermediate embedding $\boldsymbol{s_i}$ for each event $e_i$, which is later projected to obtain $\epsilon_i$. We add $\epsilon_i$ to $t_{\pi_i}  \approx  \Pb[i,:]\tb$ to ensure the CTES follows chronological order subject to inequality~\eqref{eq:AB}.  $\epsAttn$ consists of three layers.

\ehdr{Input layer}
From the first stage, we use $\mathcal{H}_{\Pb}$ to obtain trainable embedding for every event ($\zb_{\pi}$) using the mark and time as: $\zb_{\pi_i} = \wb_{c} c_{\pi_i} + \wb_{t} t_{\pi_i} \approx \wb_{c} \Pb[i,:] \bm{c} + \wb_{t} \Pb[i,:] \tb $. We also add a \textit{continuous-time} positional encoding to the event embedding~\cite{zuo2020transformer, proactive}.\\
    \ehdr{Attention layer}
We use a \textit{masked} self-attention layer to aggregate and represent $\boldsymbol{Q}$, $\boldsymbol{K}$ and $\boldsymbol{V}$ as trainable \textit{Query}, \textit{Key}, and \textit{Value} matrices respectively. Finally, we compute self-attention output as:
\begin{equation}
\boldsymbol{s}_i =  \sum_{j=1}^{i} \frac{\exp\left( (\boldsymbol{Q}  \zb_i)^T (\boldsymbol{K}  \zb_j) /\sqrt{D} \right)}{\sum_{j'=1}^{i}\exp\left( (\boldsymbol{Q}  \zb_{i})^T (\boldsymbol{K}  \zb_{j'}) /\sqrt{D} \right)} \,\boldsymbol{V}  \zb_j,
\end{equation}
where $D$ denotes the number of hidden dimensions. Here, we compute the attention weights via a softmax function and introduce the necessary non-linearity to the model to obtain the final output $\overline{\boldsymbol{s}}_i$.\\
\ehdr{Output layer}
To obtain continuous-time 
noise for the event 
$e_{{\pi}_i}$, \ie\ 
$\epsilon_i$, we project the 
hidden representation to a 
scalar value as $\epsilon_i 
\leftarrow \wb_{\epsilon} 
\overline{\boldsymbol{s}}_i + 
\boldsymbol{b}_{\epsilon}$. 
Here, $\wb_{\epsilon}, 
\boldsymbol{b}_{\epsilon}$ are 
trainable parameters. The 
total set of parameters is
$\phi=\set{\wb_{\bullet}, \boldsymbol{b}_{\bullet}, 
\boldsymbol{Q},\boldsymbol{K}, \boldsymbol{V}}$.

\xhdr{Differentiable training objective} By replacing $\pi$ with the soft permutation matrix $\Pb$ and introducing the neural networks $GS_{\theta}$ and $\epsAttn$ in the optimization~\eqref{eq:attack_final}, and aggregating the objective for all $\Hcal\in \Dcal$, we have:
\begin{align}
  & \hspace{-2mm} \underset{\theta,\phi}{\mini} \sum_{\Hcal\in \Dcal} \big(  \Lcal (\Hcal \given \Mcal_{\wb},\Hcal_{\Pb,\bm{\noise}}) + \rho_{D}  \dist(\Hcal,\Hcal_{\Pb,\bm{\noise}})\nonumber\\
  &\hspace{2cm} + \rho_{A,B}  \textstyle\sum_{i\in[|\Hcal|]}[\Ab \bm{\noise} -\Bb \Pb\tb]_+[i]  \big)\label{eq:attack_final_diffable},
  \\ 
  & \text{where, } \Pb = GS_{\theta}(\Hb) \text{ with } \Hb = \amodel(\Hcal);\ \text{ and }\ \epb\ \text{s.t.}\nonumber\\
  &\epb = \epsAttn(\Pb,\Hcal). \nn
\end{align}
In contrast to the optimization~\eqref{eq:attack_final}, which seeks to solve $\pi$ for each $\Hcal$ separately, the training of $\theta, \phi$ will provide the inductive biases of $\pi$ and $\epb$ into $GS_{\theta}$ and $\epsAttn$. With enough expressive power, they would be able to ensure $\dist(\Hcal,\Hcal_{\Pb,\epb})$ to be small and  $\Ab\epb \le \Bb \tb_{\pb}$ for every $\Hcal$ drawn from the same distribution $ \Dcal$.

\xhdr{Adversarial training for TPP} 
Given a dataset of sequences $\Dcal=\set{\Hcal}$, our adversarial training  follows the following robust optimization problem:
\begin{align}
&\hspace{-2mm}\max_{\wb}     \ \underset{\theta,\phi}{\min} \ \sum_{\Hcal\in \Dcal} \hspace{-1mm}\big(   \Lcal (\Hcal \given \Mcal_{\wb},\Hcal_{\Pb,\bm{\noise}}) + \rho_{D}  \dist(\Hcal,\Hcal_{\Pb,\bm{\noise}})\nonumber\\ 
  & \hspace{2cm} + \rho_{A,B}  \textstyle\sum_{i\in[|\Hcal|]}[\Ab \bm{\noise} -\Bb \Pb\tb]_+[i] \big).\label{eq:defense}
\end{align}

\section{Experiments} \label{exp}
Here, we evaluate \our\ by its ability to reduce the performance of learner's MTPP $\lmodel$. Specifically, we ask the following research questions.
\textbf{(RQ1)} How much performance deterioration can \our\ cause when $\lmodel$ is trained on clean examples?
\textbf{(RQ2)} How much deterioration can \our\ cause when $\lmodel$ is trained on adversarial examples?
\textbf{(RQ3)} How does \our\ perform in the context of adversarial defense? Does it provide better robustness than other adversarial strategies? All codes for \our are available at: \underline{https://github.com/data-iitd/advtpp}.

\begin{table}[t]
    \small
    \addtolength{\tabcolsep}{-2pt}
    \centering
    \begin{tabular}{l|cccc}
        \toprule
        \textbf{Dataset} &  \textbf{Taobao} & \textbf{Twitter} & \textbf{Electricity} & \textbf{Health}\\
        \hline
        \#Events $\sum_{\Hcal\in \Dcal}|\Hcal|$ & 0.5M & 2.1M & 60M & 60M\\
        \#Sequences $|\Dcal|$ & 2000 & 22000 & 10000 & 10000 \\
        \#Marks $|\Ccal|$ & 17 & 3 & 5 & 5\\
        Mean Length & 51 & 41 & 300 & 500\\
        Mark imbalance &  25.72 & 22.02 & 65.09 & 41.50 \\
        \bottomrule
    \end{tabular}
        \caption{Dataset statistics. Here, mark imbalance means the average of the ratios of the highest frequent mark and the lowest frequent mark across different $\Hcal\in\Dcal$.}
\label{tab:data-main}
\end{table}

\begin{table*}[t!]
\centering
\small
\tabcolsep 2mm 
\begin{tabular}{ l | c | c | c | c | c | c | c | c} 
 \specialrule{.12em}{.1em}{.1em} 
  \multirow{ 2}{*}{\textbf{Method}} & \multicolumn{4}{c|}{\textbf{MAE}} &\multicolumn{4}{c}{\textbf{MPA}} \\
  \cline{2-9}
& Taobao & Twitter & Electricity & Health & Taobao & Twitter & Electricity & Health \\ \hline
No Attack & 0.973 & 0.082 & 0.005 & 0.036 & 43.70 & 60.45 & 92.95 & 63.68 \\
\hline
PGD (BB) & 0.721 & 0.107 & 0.038 & 0.051 & \secondbest{19.20} & 59.81 & 92.98 & 63.59 \\
\mifg (BB) & 0.725 & 0.110 & 0.039 & 0.052 & 19.66 & 59.79 & 92.97 & 63.58 \\
\robtsd (BB) & 0.920 & 0.115 & 0.051 & 0.054 & 29.01 & 59.69 & 92.91 & 63.48 \\
\robtsp (BB) & \secondbest{1.013} &  \bestmodel{0.191} &  \bestmodel{0.063} &  \secondbest{0.056} & 36.77 & \secondbest{58.78} & \secondbest{92.86} & \secondbest{62.90} \\
\our (BB) & \bestmodel{1.383} & \secondbest{0.115} & \secondbest{0.053} &  \bestmodel{0.123} &  \bestmodel{15.97} &  \bestmodel{56.56} &  \bestmodel{91.87} &  \bestmodel{61.67} \\
\hline
\hline
PGD (WB) & 0.708 & 0.101 & 0.015 & 0.047 & \secondbest{18.43} & 59.86 & 92.90 & 63.63 \\
\mifg (WB) & 0.730 & 0.103 & 0.048 & 0.048 & 19.52 & 59.84 & 92.91 & 63.61 \\
\robtsd (WB) & 0.923 & 0.108 & \secondbest{0.072} & 0.051 & 28.94 & 59.74 & \secondbest{92.78} & 63.52 \\
\robtsp (WB) & \secondbest{1.309} & \secondbest{0.187} & 0.070 & \secondbest{0.155} & 42.52 & \secondbest{58.90} & 92.83 & \secondbest{62.87} \\
\our (WB) & \bestmodel{1.350} & \bestmodel{0.120} & \bestmodel{0.078} & \bestmodel{0.230} & \bestmodel{9.21} & \bestmodel{55.84} & \bestmodel{87.96} & \bestmodel{42.64} \\
\hline
\hline
\end{tabular}
\caption{Performance deterioration between adversarial attack strategies, where the attack is executed on the learner's MTPP model $\Mcal_{\widehat{\wb}}$ trained on clean CTESs. We use THP~\cite{zuo2020transformer} as $\Mcal_{\bullet}$ for both learner and adversary. BB and WB refer to black-box and white-box variants of the attacks. Across all cases, we ensure that 
$\dist(\Hcal,\Hcal')$ is approximately same. Higher the MAE and lower the MPA,  more successful is the attack.  \bestmodel{Bold} (\secondbest{underlined}) indicate the best and the second best model.}
\label{tab:eval_on_clean}
\end{table*}


\begin{table*}
\centering
\tabcolsep 2mm 
\small
\begin{tabular}{ l || c | c | c | c | c | c || c | c | c | c | c | c} 
     \specialrule{.12em}{.1em}{.1em}
     \centering
  &\multicolumn{6}{c||}{\textbf{Taobao}} & \multicolumn{6}{c}{\textbf{Health}}\\ \hline
       \textbf{Attack Method $\downarrow$} & \multicolumn{3}{c|}{\textbf{MAE}} &\multicolumn{3}{c||}{\textbf{MPA}} & \multicolumn{3}{c|}{\textbf{MAE}} &\multicolumn{3}{c}{\textbf{MPA}}\\
      \cline{2-13}
      \textbf{Adv. Method $\rightarrow$}  & PGD & \mifgshort & \robtsd & PGD & \mifgshort & \robtsd &    PGD & \mifgshort & \robtsd & PGD & \mifgshort & \robtsd  \\ \hline
No Attack & 0.908 & 0.905 & 0.860 & 46.31 & 46.53 & 47.09 & 0.034 & 0.034 & 0.033 & 64.37 & 64.36 & 64.34\\ \hline
PGD (BB) & 0.666 & 0.678 & 0.712 & \secondbest{42.09} & \secondbest{42.21} & \secondbest{44.07} & 0.034 & 0.034 & 0.034 & 64.40 & 64.38 & 64.36\\
\mifg (BB) & 0.671 & 0.682 & 0.718 & 42.22 & 42.37 & 44.14 & 0.034 & 0.034 & 0.034 & 64.40 & 64.39 & 64.36\\
\robtsd (BB) & 0.812 & 0.861 & 0.881 & 43.38 & 44.32 & 45.54 & 0.034 & 0.034 & 0.033 & 64.39 & 64.38 & 64.35\\
\robtsp (BB) & \secondbest{0.941} & \secondbest{0.937} & \secondbest{0.923} & 46.64 & 46.73 & 47.81 & \secondbest{0.036} & \bestmodel{0.035} & \secondbest{0.034} & \secondbest{64.31} & \secondbest{64.36} & \secondbest{64.29}\\
\our (BB) & \bestmodel{1.163} & \bestmodel{1.138} & \bestmodel{1.061} & \bestmodel{41.52} & \bestmodel{41.42} & \bestmodel{41.58} & \bestmodel{0.049} & \secondbest{0.034} & \bestmodel{0.035} & \bestmodel{49.83} & \bestmodel{61.65} & \bestmodel{60.29}\\  
    \hline \hline
PGD (WB) & 0.670 & 0.681 & 0.717 & \secondbest{42.00} & \secondbest{42.12} & \secondbest{43.97} & 0.034 & 0.034 & 0.034 & 64.39 & 64.39 & 64.36\\ 
\mifg (WB) & 0.674 & 0.686 & 0.723 & 42.17 & 42.26 & 44.06 & 0.034 & 0.034 & 0.034 & 64.39 & 64.39 & 64.36\\
\robtsd (WB) & 0.816 & 0.864 & 0.884 & 43.56 & 44.47 & 45.68 & 0.034 & 0.034 & 0.033 & 64.39 & 64.39 & 64.35\\
\robtsp (WB) & \secondbest{1.147} & \secondbest{1.138} & \secondbest{1.060} & 47.31 & 47.50 & 48.00 & \secondbest{0.034} & \secondbest{0.035} & \secondbest{0.034} & \secondbest{64.29} & \secondbest{64.11} & \secondbest{64.30}\\
\our (WB) & \bestmodel{1.265} & \bestmodel{1.295} & \bestmodel{1.228} & \bestmodel{18.56} & \bestmodel{19.94} & \bestmodel{21.12} & \bestmodel{0.042} & \bestmodel{0.048} & \bestmodel{0.041} & \bestmodel{40.80} & \bestmodel{53.67} & \bestmodel{49.14}\\
    \hline\hline
    \end{tabular}
\caption{Performance on different adversarial attack strategies, where the attack is executed on an MTPP model $\Mcal_{\widehat{\wb}}$ learned using adversarial training methods -- PGD, \mifg (MIF) and \robtsd. We ensure that $\dist(\Hcal,\Hcal')$ is approximately same.}
\label{tab:eval_exp2}
\end{table*} 

\xhdr{Datasets}
We use Taobao~\cite{taobao}, Twitter~\cite{MeiE16}, Electricity~\cite{refit}, and Health~\cite{neuroseqret, ecg} datasets. We provide the details for each dataset in Table~\ref{tab:data-main}.

\xhdr{Baselines} In our knowledge, \our is the first adversarial attack on the MTPPs. As a result, we resort to the adversarial methods designed specifically for other domains and adapt them for MTPPs. 
We use two attack strategies from the domain of computer vision, \viz, (1) PGD~\cite{madry},
(2) \mifg~\cite{dong2018boosting}; 
and two attack strategies which were 
used for discrete time-series models, \viz,
(3) \robtsd\ and
(4) \robtsp.~\cite{liu2023}. 


\xhdr{White box and black box models}
Depending on how much information about the learner's model $\lmodel$ is available to the adversary, the adversary can use two types of $\amodel$: 
(1) Black box (BB): Here, $\amodel \neq \lmodel$, \ie, the adversary uses another version of the MTPP model $\Mcal _{\wb}$ that is not same as same as the learner's model $\Mcal _{\widehat{\wb}}$. There can be several examples of such black box models $\Mcal _{\wb}$. Predominantly, we assume that the adversary uses a model which is trained on the same training set with different seed. 
(2) White box (WB): Here,
 $\Mcal _{\wb} = \Mcal _{\widehat{\wb}}$, \ie,  the adversary has access to the learner's model, which is going to be deployed during test; and the adversary uses this exact trained model to compute the perturbation. 
 
 \xhdr{Evaluation Metrics}
Given a dataset of CTESs $\Dcal=\set{\Hcal}$, we split $\Dcal$ into 70\% training, 10\% validation and 20\% test set.
Given   the adversary's model $\Mcal _{\wb}$,  we  compute the adversarial perturbations by solving the optimization problem~\eqref{eq:attack_final_diffable}  for our method.
Similarly, for baseline methods, we compute the worst case perturbations, by minimizing the log-likelihood using their respective algorithms.
Then, we feed the perturbed test CTESs to the learner's model $\lmodel$ and report the results   
in terms of  
mean absolute error (MAE) between predicted and current time and mark prediction accuracy (MPA) 
to assess time and mark prediction ability respectively. Mathematically, $\mathrm{MPA} = \frac{1}{|\Hcal|}\sum_{e_i\in \Hcal} \mathbb{I}(c_i=\widehat{c}_i)$ and $\mathrm{MAE} = \frac{1}{|\Hcal|}\sum_{e_i\in \Hcal} [|t_i-\widehat{t}_i|]$. For the evaluation of adversarial attacks on the same learner's model $\Mcal_{\widehat{\wb}}$, a lower MPA and higher MAE correspond to a better attack framework, given that all other conditions are fixed, whereas, for the evaluation of defences a higher MPA and lower MAE correspond to a better defense framework.

\subsection{Experimental Results}

\xhdr{Effect of adversarial attacks on models trained on clean examples}
Here, we compare the extent of the performance degradation of the  learner model $\lmodel$, caused by \our\ against the baselines, 
when $\lmodel$ is trained on clean CTESs (\textbf{RQ1}).
We observed that the  adversarial perturbation $\dist(\Hcal,\Hcal')$ is 
extremely difficult to control for the baselines, since there is no explicit ``knob'' in their model, which can regulate $\dist(\Hcal,\Hcal')$. As a result, we could not equalize this distance across different methods. However, we ensured that this distance for all the adversarial attack strategies remains within 10\% deviation.  
We present the results in
Table~\ref{tab:eval_on_clean}. It shows that \our degrades the predictive performance of the learner's model $\lmodel$, significantly compared to the baselines, for fourteen out of sixteen cases. 
\textbf{In particular, for Electricity dataset, only we achieve less than 90\% MPA performance. In Taobao, \our\ outperforms the second best baseline by 36.5\% MAE and 16.8\% MPA for black box attack and by 3\% MAE and 50\% MPA in the white-box setting.
}

\begin{figure}[t!]
\centering
\includegraphics[width=\columnwidth]{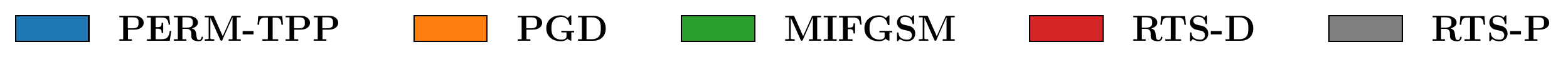} 
\begin{subfigure}{\linewidth}
  \centering
  \includegraphics[height=2.0cm]{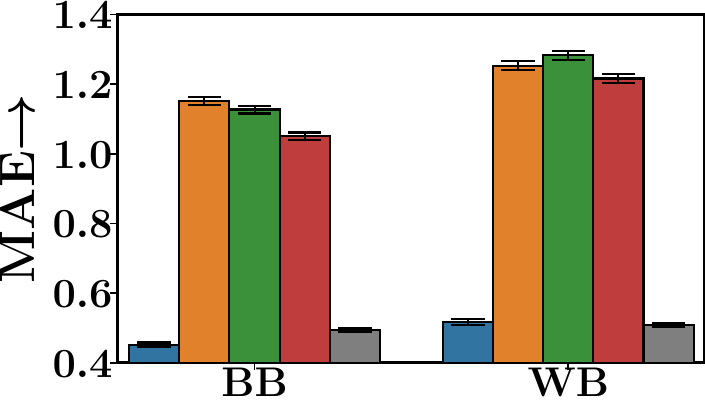}
  \hspace{0.5cm}
  \includegraphics[height=2.0cm]{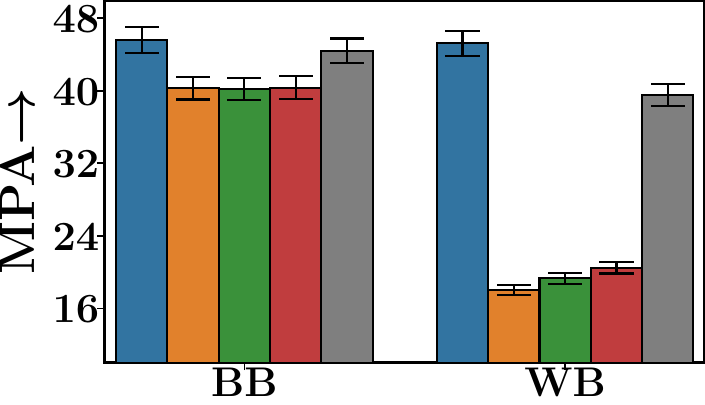}
\end{subfigure}
\caption{Adversarial Robustness of $\lmodel$, trained using different adversarial training methods and \our\ attack for Taobao dataset. Here, lower the MAE and higher the MPA, more successful is the defense model.}
\label{fig:eval_exp2_2}
\end{figure}

\xhdr{Effect of adversarial attacks on models trained on adversarial examples}
Here, we address \textbf{RQ2} by evaluating how successfully 
\our\ can deteriorate the performance of the learner's model $\lmodel$, when it is trained on adversarially perturbed sequences. 
 Table~\ref{tab:eval_exp2} shows the results across two datasets, for three adversarial training methods:  PGD, \mifg\ and \robtsd, which were used to train $\lmodel$.  We observe that 
(1) \our degrades the attacks more severely than the baselines in all cases, except for MAE in black-box setting for the Health dataset, when using the \mifg adversarial training setup. 
(2) In many cases, adversarial training improved predictive performance under 
adversarial attack, from training on clean sequences (Table~\ref{tab:eval_on_clean} vs Table~\ref{tab:eval_exp2}).

\xhdr{Adversarial defense}
Here, we address \textbf{RQ3} by comparing the adversarial robustness of  the learner's model $\lmodel$, across different adversarial training methods. 
Figure~\ref{fig:eval_exp2_2} shows the results for \our\ attack in Taobao dataset, which highlight that: (1) \our achieves the highest resilience towards attacks in terms of MPA; (2) For MAE, \our performs competitively with \robtsp; and wins over \robtsp\ for black box setting.

\section{Conclusion}
We propose a trainable adversarial attack specifically designed for MTPP. Our novel two-stage attack strategy, using differentiable neural surrogates for permutation and noise generation, gives us enhanced degradation capabilities on diverse datasets. \our opens up several future directions, for e.g., generating differentially private CTES examples. 

\section*{Acknowledgements}
Srikanta acknowledges DS Chair of AI Fellowship, Abir acknowledges a Trust Lab Grant from IITB.

\bibliography{references}
\clearpage

\end{document}